**JCSTS**

AL-KINDI CENTER FOR RESEARCH
AND DEVELOPMENT

| RESEARCH ARTICLE

# An Approach for Detection of Entities in Dynamic Media Contents

## Mbongo Nzakiese[1] ✉ and Ngombo Armando[2]

[1]Department of Informatics Engineering, Instituto de Tecnologia de Informação e Comunicação, Universidade de Luanda, Luanda, Angola

[2]Department of Research, Innovation, Entrepreneurship & Post-graduate activities, Universidade Rainha Njinga a Mbande, Malanje-Angola; Department of Informatics Engineering, Instituto Politécnico da Universidade Kimpa Vita, Uíge, Angola

**Corresponding Author:** Mbongo Nzakiese, **E-mail:** arlainnzakiese@gmail.com

| ABSTRACT

The notion of learning underlies almost every evolution of Intelligent Agents. In this paper, we present an approach for searching and detecting a given entity in a video sequence. Specifically, we study how the deep learning technique by artificial neural networks allows us to detect a character in a video sequence. The technique of detecting a character in a video is a complex field of study, considering the multitude of objects present in the data under analysis. From the results obtained, we highlight the following, compared to state of the art: In our approach, within the field of Computer Vision, the structuring of supervised learning algorithms allowed us to achieve several successes from simple characteristics of the target character. Our results demonstrate that is new approach allows us to locate, in an efficient way, wanted individuals from a private or public image base. For the case of Angola, the classifier we propose opens the possibility of reinforcing the national security system based on the database of target individuals (disappeared, criminals, etc.) and the video sequences of the Integrated Public Security Centre (CISP).

| KEYWORDS

Entity Detection, Computer Vision, Multi-instance, Public Safety

| ARTICLE INFORMATION

## 1. Introduction

One of the fundamental abilities of human beings is to analyse their environment. In most cases, this involves recognising elements in our field of vision: finding other people and identifying cars and animals. In this paper, we explore the concept of Computer Vision as an approach to detecting an individual in a video sequence. Computational Vision (VC) is a branch of Artificial Intelligence (AI) which aims to modelling and replicate human vision with a computer system (Sultani, Chen & Shah, 2018). For this purpose, Computational Vision "VC" studies how to analyze, understand, reconstruct, interrupt, and detect properties or features of a scene in n dimensions (nD) from its images in two dimensions (2D). Thus, we consider that VC is a way through which an Intelligent Agent can understand what it observes by means of an electronic vision captured by film cameras. Ultimately, the environment observed in these electronic devices is composed of sequences of images (called frames), which represent the activities and changes of states of the captured environment[2]. In the structuring of a video, the frames parade at a certain frequency, measured in frames per second (FPS) or Frames per second in Portuguese. Most video surveillance cameras have FPS 16 and 48, that is, 16 frames per second and 48 frames per second. The cameras of today's smartphones have a frequency that varies between 24 and 240, and finally, today's professional cameras have frequencies of up to 10 billion FPS, which allows them to capture the path of light, film transparent objects and very ephemeral phenomena such as a shock wave or the signals transmitted by neurons.

One of the research challenges has been the possibility to extract, analyse and understand the information contained in a frame. This computer vision and image recognition by machines has been possible thanks to algorithmic techniques used in Machine

Learning and Deep Learning. VC and image recognition are terms often used as synonyms. However, the former covers more actions than those of image analysis and perception. As the ways to analyse and identify moving objects in videos become more widespread, we will continue to see the emergence of applications with higher accuracy in environments or systems where vision understanding is indispensable to analyse the existence, absence or change of state of a given environment. These environments include, for example, automotive security systems, airport security systems, surgical interventions, and fraud detection systems.

For various purposes competing to affirm their prosperity, various organizations need to extract information from the states of the environments that have been captured by the cameras. Now, the more massive and diverse the data captured by the cameras, the more complex the process of detecting a particular object that may represent the expected information for the organization. The detection process is very broad in terms of objects. For example, we can apply it to human feelings and animal states through their voices and emitted sounds, respectively. We also have object detection in photographs and the detection of handwritten characters. In the case of this study, the focus is on character detection within dynamic video sequences, i.e., both the objects to be detected and the image capture angles are in motion.

This paper is structured as follows. The second section reviews the related works, followed by the approach selected for the study. In the fourth section, we have the implementation and tests done. In the fifth section, we will present the essentials of the results obtained before concluding with the sixth section.

## 2. Literature Review

The main modules for the detection and recognition of actions, people, faces, cars, and accidents involve the detection and classification of objects, pattern recognition and the understanding of what has been detected. As an initial study in the detection context, in it is shown how surveillance cameras can capture a variety of anomalies in real scenarios, and the authors propose machine learning techniques to detect such situations by exploring normal and anomalous videos (Sultani, Chen & Shah, 2018).

In (Bertolino, Foret, & Pellerin, 2001), the authors present a movement segmentation algorithm to detect people who move in indoor environments. The algorithm proposed by them leverages moving camera sequences and is composed of two main parts. In the first one, a frame-by-frame procedure is applied to calculate the difference image and a neural network is used to classify whether the resulting image represents a static scene or a scene containing moving objects. The second part tries to reduce detection errors in terms of false trues. A finite state automaton is designed to provide robust classification and reduce the number of false or lost drops. Looking at two typical cases, an empty scene and a scene with a moving object, the problem we want to deal with becomes clear: thresholding techniques can classify acquisition noise as moving pixels. To solve this problem, the study developed in (Forestic. Michelonic & Piciarelli, 2005) comes up with a new technique to correctly classify the current video sequence.

Finally, (Ahmad, Ahmed & Adnan, 2019), a new method was presented to detect and launch alarms related to the action of pedestrian falls, using, first, the frame difference method and the pedestrian markers frame. The frame difference method and the background difference method are combined to mark the destination contour, and the contour data is analyzed to correct the location of pedestrians. Then, the position of the movement, the trajectory of the pedestrian centroid and the length/width ratio of the pedestrian marker structure and the dwell time are analyzed to detect the action of falling pedestrians and other abnormal movements that finally, serve as input to trigger a buzzer.

In this way, we conclude that the approaches presented in the related works allowed us to understand the different steps necessary to carry out a system for the detection and recognition of activities in dynamic environments. This literature review also allowed us to select the desired model for the implementation of our approach, which will be presented in the next section.

## 3. Methodology

The methodology we adopted is based on the article (Sultani, Chen & Shah, 2018). For our study, we created two folders, which we also call "bag", containing several video sequences. We have a positive bag and a negative bag(Jagadeesh & Patil, 2016). A bag is negative if and only if all its instances are negative (that is, it contains no person to detect), and it is positive if at least one of its instances is positive (that is, it contains the person to detect). Therefore, the first bag contains the videos that integrate the identity to be detected, thus a specific person, while the second bag contains images from various angles of the person of interest (identity to be detected). In terms of machine learning, the goal is to maximize the separation between the two bags to obtain a model capable of producing different results as the person of interest appears (or does not) in a given video segment. In the next section, we describe, in more detail, the dataset used.

$$X_m = \{x_1, x_2, \ldots, x_n\} \tag{1}$$

Expression (1) summarises the organisation of the video segments belonging to a bag, where xi is the i-th bag and m is the number of instances in each bag.

### 3.1 Multi-instance learning – MIL

In supervised learning, each instance is described by a feature vector and associated with a class label that identifies which object family it belongs to. In MIL, which generalizes supervised learning, each instance is also described by a feature vector. However, in our case, the class tag is not associated with an instance but rather associated with the bags to which the images or videos belong. In expression (2), we consider the training dataset B = {B1, B2, . . ., BZ} where Z is a set of bags, each bag is associated with a tag Yi ∈ {−1, 1} and that it contains Ni instances; then Bi = {xi1, xi2, . . ., xiNi} an instance Xij can be considered to be labeled positive or negative depending on the bag in which it belongs to, so we say that yij ∈ {− 1, 1} (J. Amores, 2013).

$$Yi = \begin{cases} +1, if \; \exists y \in Bi: yij = +1 \\ -1, if \; \forall y \in Bi: yij = -1 \end{cases} \tag{2}$$

### 3.2 Optimization

For the optimization, the cost function of Hinge Loss was used, which allows classifying the model with the values {1,1}. The output layer of an SVM network was configured to have a single neuron with a hyperbolic tangent activation function capable of generating a single value in the range of [1,1] (Dietterich, Lathrop & LozanoPérez, 1997). The following expression (3) is a minimization of the Hinge function, where z is the total number of bags containing training examples, the element in parentheses is the cost function, YBi represents the bag-level label, and Bp represents a positive video as a positive bag where the temp segments create positive individual instances (p1, p2, …pm). In (3), Bn denotes the negative label and (n1, n2, …nm), φ(x) designates the representation of resources of a patch image or video segment. Finally, B is a bias, and w is the classifier (Quddus, Fieguth & Basir, 2005; Krizhevsky, Sutskever & Hinton, 2012).

$$\min_{w} \frac{1}{z} \sum_{j=1}^{z} \left( max \left( 0,1 \; - Y_{B_j} \left( \max_{i \in B_j} (w.\phi(X_i)) - b \right) \right) \right) + \frac{1}{2} \parallel \mathcal{W} \parallel^2 \tag{3}$$

In the field of human character detection in cartoon videos, it is difficult to capture effective and discriminative features as a result of body variations and style changes. The variations are mainly caused by scale, clothing, postures, back, front or side position; camera aim fixation and scene dynamism. For Wan et al., when the input of the classification system is a video sequence instead of a static image, the detection task requires more effort since the model needs to learn from temporal features. In the process of deep learning, features can be automatically learned from temporal and spatial domains simultaneously (Lin, Tiancheng et al., 2023).

Figure 2 is a flowchart that illustrates how our learning process is done. We have in inputs a dataset with video sequences of different temporal lengths, which size will then be uniformed to 32 frames per second. We thus obtain a temporal video segment and place the positive and negative videos in the input of the C3D network. Next, we obtain at the output of the C3D network the features because of the extraction. The extracted features of this three-dimensional convolutional network in fc61 file format are placed on the input of our neural network, which is of Fully-Connected type, to obtain the trained model for An Approach for the detection of entities in dynamic media contents.

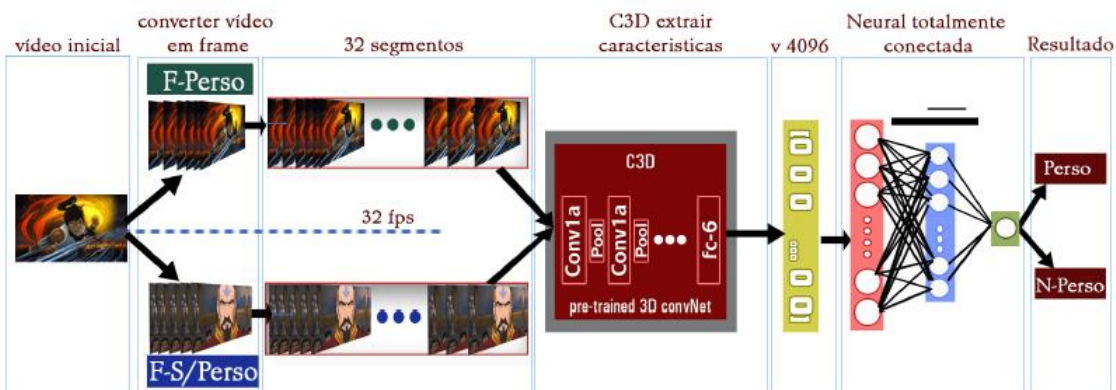

**Figure 1:** Flowchart of the proposed model for Supervised learning framework in multimedia search in the case of detecting a character in cartoon video sequences. (Adapted from (Sultani, Chen & Shah, 2018)).

In multi-instance regression and also in Figure 1, each bag is associated with a single real number, as in a standard regression approach. In Multiple Instance 'MI' regression, it is considered that there is one instance in each bag called the "main instance", which determines the bag's label (Lin, Tiancheng et al., 2023). The ideal goal of MI regression is to find a Hyper plane that minimizes the squared cost function of the primary instances in each bag.

One of the goals of our implementation is that the video segments of positive bags (videos with characters to identify) have higher identification or detection scores than the video segments of negative bags, where the target entity is not found. To do this, the straightforward approach would be to use a classification cost function to obtain in the model output high scores on positive video segments (of the character to detect) and penalise the scores of the segments on negative videos. However, since the video segments we have are not annotated, this is where we propose the objective function for the classification of the Multiple instances. The objective function represents the goal to be achieved for the MI problem. The set of constraints defines conditions on the state space that the variables must satisfy. These constraints are often inequality or equality constraints and generally limit the search space (feasible solutions) proposed by https://en.wikipedia.org/wiki/Multiple_instance_learning.

The optimal solution to this problem is to find the point or set of points in the search space that best satisfies the objective function. The result is called the "optimal value". The main goal while learning our model is that the Loss de "hinge" is minimized, also preventing the model from falling into overfitting. By training many positive and negative bags, we expect our model to be as generalized as possible to allow all positive bags to have higher scores while negative segments have the lowest possible scores.

### 3.3 Implementation and testing
In the previous chapter, we described our model, presenting the context in which it works. In this chapter, we present the experiment carried out with the model and the results obtained.

### 3.3.1 Data set description
The data set used contains a large variety of brightness, colours, spaces and backgrounds that are not uniform. These data, in most cases, are raw and require treatment before use and how they will be used. The data set is separated into two groups, namely: data for model training (80%) and data for model testing and validation (20%).

### 3.4 Data collection(videos)
The dataset consists of video sequences of cartoons of the avatar legend "korra". We obtained the videos of the series (seasons) in a total of almost 20 hours and 20 minutes, as described in Table 1. In most of the video sequences available in other sources such as the YouTube platform, the scenarios were presented in a very repetitive and mixed way, downloaded at http://www.nick.co.uk/shows/legendofkorra/videos/airbendingbasics/0sxstw as described in Table.

**Table 1: Description of videos by books (series or season)**

| Books | Number of videos per series | Duration min - Duration max | Book duration |
|---|---|---|---|
| Book_1_Ar | 12 | 22'40" - 25':25" | 287' |
| Book_2_Espirito | 14 | 23':02" - 23':57" | 333' |
| Book_3_Mundaça | 13 | 22':02" - 22':58" | 294' |
| Book_4_Equilíbrio | 13 | 22':48" - 23':00" | 928' |
| - | - | - | - |
| Total | 52 | 22':02" - 25":25" | 1212' |

The first column of Table 2 represents the set of seasons, and each row in this column represents a season of the Legend, namely Book 1 Ar with 12 episodes, where the video with the shortest duration has 22 minutes and 40 seconds, and the video with the longest duration has 25 minutes and 25 seconds. Thus, the total duration of Book 1 Ar is 287 minutes. In the same vein, Book 2 Spirit contains 14 episodes, where the video with the shortest duration is 23 minutes and 02 seconds, and the video with the longest duration is 23 minutes and 57 seconds, so the total duration of Book 2 Spirit is 333 minutes. Book 3 Change contains 13 episodes, where the video with the shortest duration counts 22 minutes and 02 seconds, and the video with the longest duration counts 22 minutes and 58 seconds, so the total duration of the book Book 3 Change is 294 minutes. Book 4 Balance contains 13 episodes, where the video with the shortest duration counts 22 minutes and 48 seconds, and the video with the longest duration is 23 minutes and 00 seconds; thus, the total duration of Book 4 Balance is 298 minutes.

### 3.5 Organisation and Notation
To organise our dataset composed of 52 video sequences, we first eliminated the audio component from them as it is not an irrelevant feature in the detection phase. Next, we converted the videos into frames (images). The frames from a given video were grouped in a single folder so that each folder represents a video sequence whose names were made in order of frame extraction. Furthermore, we eliminated the parts that do not provide any useful information, that is, in which the features extracted from the sequences were not part of the features explored by the model, such as the writings and comments at the beginning and end of the video. As already stated in the points above and the previous chapter, we did not classify the videos in detail into classes. The only way to annotate the videos was to put all the positive videos in a single folder and those that were negative in another folder(Yang et al. 2006; Carneiro et al. 2007).

### 3.6 Resizing videos

As the videos were taken from two different sources, they were not the same size in terms of resolution. The first source provided videos with a size of 640 x 360 pixels, and the second one was 1920 x 1080 pixels. Therefore, all the images in our dataset were resized to 640 x 360 pixels for better extraction of the C3D features and model testing. The normalisation consisted in reconstructing the temporary videos consisting of 32 FPS from the frames that we considered important and that we annotated " X" in the dataset organisation process (Carneiro et al., 2007). Therefore, each video finally had a duration of 16 seconds, equivalent to 480 frames. Finally, we separated the videos that contained characters (videos labelled positive) we placed in a folder and those that did not have characters (negative videos) in another folder, as illustrated in Figure 5 (Kim, Jaehwan et al. 2017).

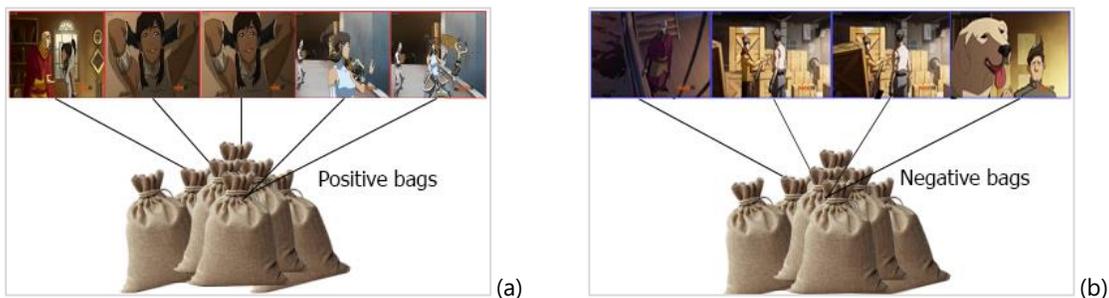

(a)                                                                                                    (b)

**Figure 2:** Video sequences for target entity detection. (a) Positive case - with the target entity inside the folder. (b) Positive case - without the target entity inside the folder

Firstly, the error we avoided was that of assessing the quality of our model using the same data that would serve for the training phase. In fact, the dataset for training is a set of examples used to adjust the parameters of the model that is fully optimised for the data with which it was trained. It was, therefore, necessary to evaluate the model by a set of data unfamiliar to the model.

**Tabela 2:** Divisão do conjunto de dados

| Dataset | Number of videos | With Avatar Korra | Other people |
|---------|------------------|-------------------|--------------|
| Training | **3225** | **953** | **2272** |
| Test | **806** | **238** | **5682** |
| - | **-** | **-** | **-** |
| Total | **4061** | **1191** | **2840** |

Table 2 exposes the breakdown of the dataset into the training and test classes defined after the normalization process; we had a total of 4061 videos, of which 1191 were videos with the Avatar Korra characters, and 2840 were normal videos without the Avatar Korra characters. In other words, we had 1191 positive videos and 2840 negative videos, and we put 80% of the data in the dedicated folder for training and 20% for the testing phase. The data set for the training holds 953 videos with the Avatar Korra character and 2272 videos without the Avatar Korra character, and the set for the test consisted of 238 videos with the Avatar Korra character against 568 videos without the Avatar Korra character.

The experiments were performed on an i7 machine with the following other configurations:

- GPU GeForce Serial 10 (GTX1080), with an 8GO video memory;
- 16GO RAM ;
- Linux 18.04 de 64 bits;
- CUDA 10.1;
- cuDNN 7.6.5
- Python 3.7;
- PyCharm 2019.3

### 3.7 Implementation

The implementation started right after we had a ready data set; after normalization, we extracted the features. These features of the images determine or reveal the different common points (values) that exist between different images and also the different points that do not exist between them. The paper shows how we used the C3D model for feature extraction from the videos (training and testing). We extracted features from the C3D model developed by the company Meta (ex-Facebook) that we downloaded from https://github.com/facebookarchive/C3D. As we mentioned in Figure 4, we put in the input of the C3D network our dataset that contains all temporal videos of 16 seconds, and we obtained in the output one of the C3D model with a vector of 4096 dimensions. The C3D Model extracting the features for each tiny clip of 16 frames, the total of the clips extracted is 120900, knowing that one clip equals 16 frames and the extraction is done with a batch size of 50 blocks of 16 frames. Explicitly, a video is equal to a segment, so a block of 16 frames of a video represents, in this case, each instance of a segment. In the output of the model, we get a feature (characteristic) in the form of a vector of 4096 numbers that represent a segment of our dataset for the detection of the character Avatar Korra.

At https://docs.google.com/document/d/1-QqZ3JHd76JfimY4QKqOojcEaf5g3JS0lNh-FHTxLag/edit, the output of the C3D model will be used to represent the complex features of a given video. This feature will serve in the input of our Fully connected (FC) architecture classifier with 3 fully connected layers with the following parameters: 512 neurons in the first fully connected layer, with 32 neurons in the second layer and these 32 are fully connected with 1 neuron of the last layer. To avoid the system falling into overfitting too early before the learning process completes, we used a DropOut of 0.6(60%) to regularize the model. We then used the glorot_normal weight initializer where in the first layer, we used the Relu activation function, and in the last layer, we used Sigmoid (KRIZHEVSKY, SUTSKEVER & HINTON, 2012).

### 3.8 Optimization

We trained the network by optimising the classification tasks. In this experiment, we compared 4 different optimisers, and after the creation of the 4 models, we tested the effectiveness of each of them separately. It could be found that among the 4 optimisers, the accuracy of the Stochastic Gradient Descent (SGD) optimiser is relatively higher. Therefore, it was for this reason that we adopted this optimiser in the subsequent model tests. During the training process, we found that the Adam and Adagrad optimisers have the fastest convergence speed and can even save training time compared to SGD. Therefore, the results obtained after testing are very close, and the differences in relative accuracy are only visible after the commas (Quddus, Fieguth & Basir, 2005). Table 3 presents our experimental results.

**Table 3**: comparison between the optimisers tested.

| Optimiser | AUC(%) |
|-----------|--------|
| RMSprop | **66,00%** |
| Adagrad | **67,00%** |
| Adam | **67,26%** |
| SGD | **67,68%** |

The main objective of a system based on An Approach for the detection of entities in dynamic media contents in its use case of detecting a particular person, in this case, Avatar Korra, is to have the ability to distinguish Avatar Korra from the remaining unknown or unimportant people in our search method. A model system would give the right answers during the detection process, so to evaluate the performance of our model, it was inevitable that the model would check the 4 fundamental metrics:

- True Positive Rate: The pattern is true because it has Avatar Korra in the video sequences, and the classifier classifies it as such.
- Negative True Rate: The pattern is false because it does not have Avatar Korra in the video sequences, and the classifier classifies it as such.
- False Positive Rate: the pattern is False because it does not have Avatar Korra in the video sequences, but the classifier ranks it true as if it had Avatar Korra in the video sequences.
- False Negative Rate: the pattern is True because it has Avatar Korra in the video sequences, but the classifier rates it False as not having Avatar Korra in the video sequences.

**Table 4**: Confusion matrix

|  |  | Real Class | |
|---|---|---|---|
|  |  | **Positive** | **Negative** |
| Classifier class | **Positive** | TP | FN |
|  | **Negative** | FP | TN |

To evaluate our model trained with Machine Learning techniques, we used the Operating Characteristic (ROC) curve, as it shows the performance of a classification model and its limits to classification. However, its true Positive Rate, as well as False Positive Rate, is required to visualize the ROC curve (DHIMAN & VISHWAKARMA, 2019).

True Positive Rate

$$TPR = \frac{TP}{P} = \frac{TP}{TP+FN} = 1 - FNR \qquad (4)$$

False Positive Rate

$$FPR = \frac{FP}{N} = \frac{FP}{FP+TN} = 1 - TNR \qquad (5)$$

True Negative Rate

$$TNR = \frac{TN}{N} = \frac{TN}{TN+FP} = 1 - FPR \qquad (6)$$

False Negative Rate

$$FPR = \frac{FN}{P} = \frac{FN}{FN+TP} = 1 - TPR \qquad (7)$$

The standard accuracy and the area under the curve (AUC) are used as metrics for Classification. In binary classification, standard accuracy can be written as:

$$\text{Accuracy} = \frac{TP+TN}{P+N} \qquad (8)$$

$$\text{Accuracy} = \frac{TP+TN}{TP+TN+FP+FN} \qquad (9)$$

The ROC Curve for model classification.

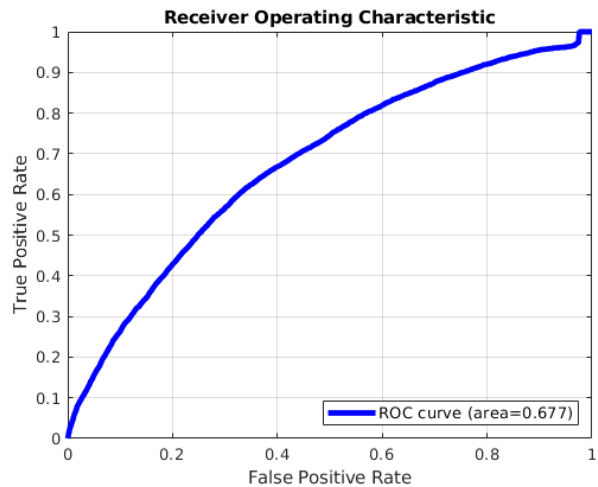

**Figure 3:** ROC curve obtained using the SGD optimiser with the ROC curve area of 67.7%.

As clarified in the points above that the dataset used in our research holds many variations; the AUC of the ROC curve that we obtain from the model trained with the optimization in SGD exceeds the areas of the ROC curve that the models trained with the networks that used optimizers other than SGD. With SGD, we achieve 67.7 of the ROC curve area (AUC).

In the next section, we present the results of applying the method we proposed by applying it to video sequences. The goal is to show the cases where our model was successful and the cases where it failed to obtain the expected results.

## 4. Results/Findings
### 4.1 Successful cases

In all the cases in which the trained model was successful, that is, in the sequences of the videos that contain the presence of the object to be detected and it was highly detected by our model. In another part of the sequences of the videos, where the model said that the person which we searched was not in the video, and it is true that she was not present in any of the sequences of the video.

Figure 4 (a) shows us Avatar Korra inside a car, and the model was able to detect it even though it was presented in profile. The sensitivity certainty during classification was at least 88%. In Figure 4 (b), the model was able to detect Avatar Korra being in the forward posture as illustrated in image 1 and graph 2, and in image 3 with graph 4, the model detected Avatar Korra even though he was on his back in the same sequences of a video. The system was able to classify both positions with a sensitivity higher than 87%. Figure 4 (c) shows us something very different than usual about Avatar Korra. The model classified her as being in profile with an outfit different from those the model had already detected in the previous figures. It is the same garment with a different colour than the previous garments.

Another example is illustrated in Figure 5 (d), where we demonstrate the effectiveness of the chosen method. The model classifies Avatar Korra, this time covering a part of her head with a hood that is very different from the clothing worn in the previous figures. Thus, the system demonstrates that although the shape of her hair is also one of the features that identify her, the model searches for many other important features that it classifies as such. During a maximum period of 15 seconds, the sequences of the same video were changed, thus capturing in different frequencies and postures the filmed characters. Now, what Figure 6 (e) shows is that despite all these changes in the camera's perception, the model was able to detect with a sensitivity of more than 93%, maintaining the specificity of the detection process above this same percentage, knowing that Avatar Korra is in a different hairstyle than usual and with her hair cut.

Finally, in Figure 7 (f), we present the success case in which the system did not identify Avatar Korra. In reality, neither she nor the features that identify her were not found in the different video sequences. The training model was trained to identify the characteristics of Avatar Korra and classify her as a detected entity, but in the images that are represented in the following figure, she (Avatar Korra) was not detected, and the model then told the truth.

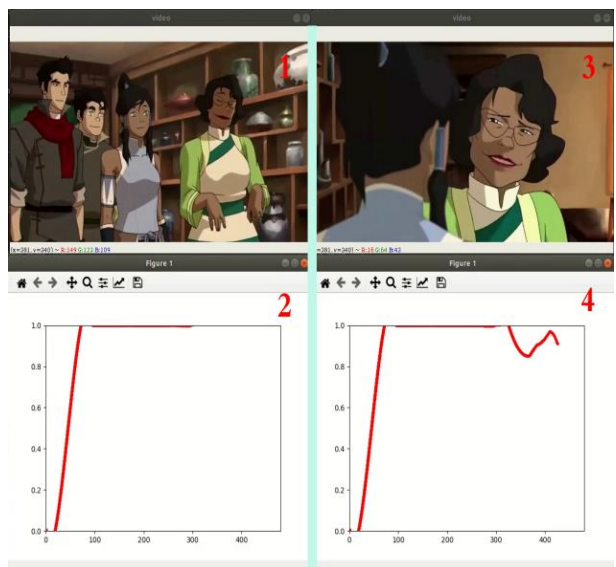

**Figure 4**: Success stories: ranking of Avatar Korra in different video sequences (image (1,2) and graph(2,4).

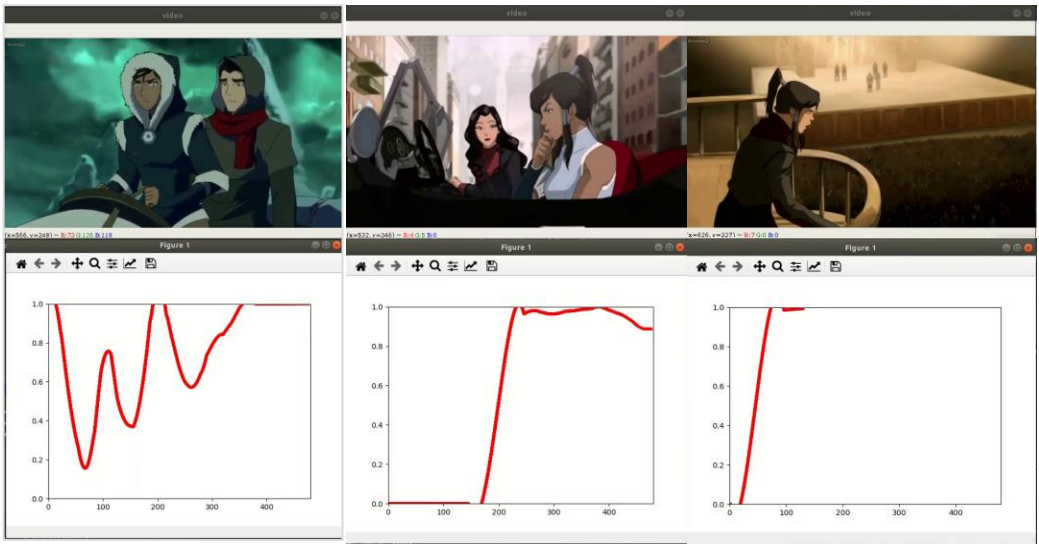

**Figure 5:** Success stories: classification of Avatar Korra in different video sequences (image (d)).

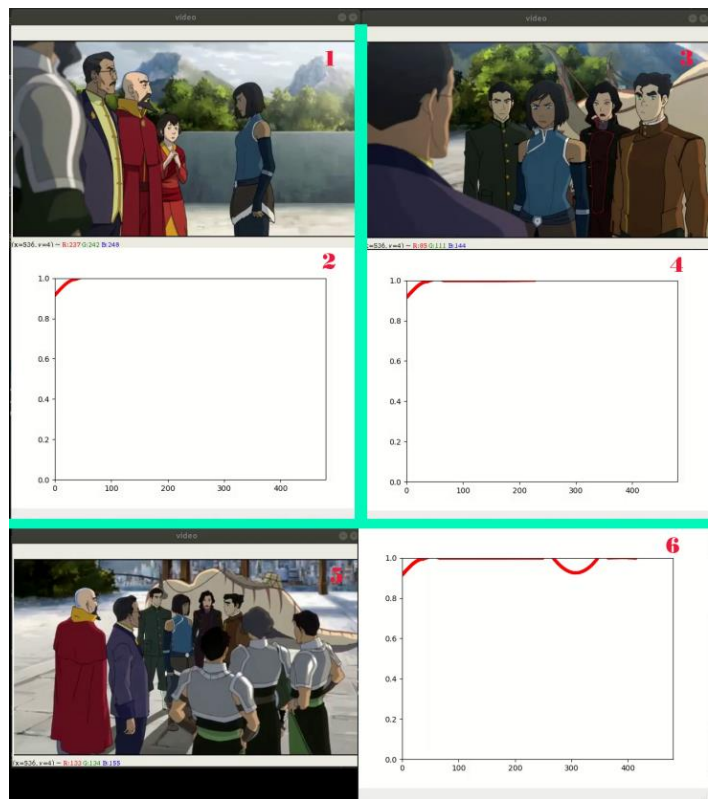

**Figure 6:** Success stories: classification of Avatar Korra in different video sequences (image (e)).

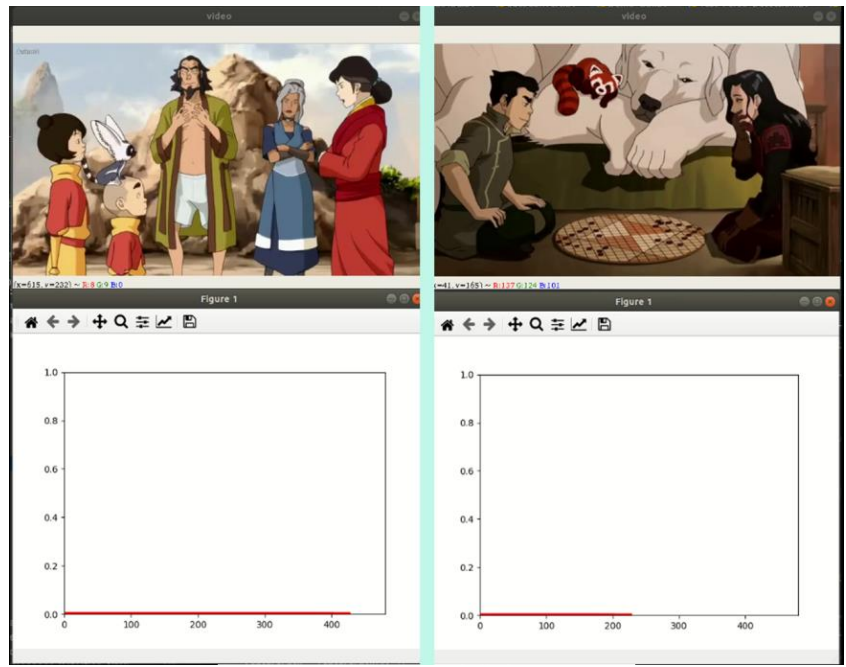

**Figure 7:** Success story: Non-rating of Avatar Korra in different video sequences (image (e)).

### 4.2 Flaws in the model

Here we present the flaws that there were in the proposed model. The dataset was one of the most complex sets as it comprised mixed image sequences with intense physical activities, facial expression (smile), change of clothing and hairstyle, change of background etc.

In Figure 8 (a), we noticed that the further away from the camera the character is, the more difficult it is to detect by the model. As soon as the character approaches the camera, the model takes a few milliseconds to detect it. In Figure 8 (b), we have a false detection alert for Avatar Korra. However, observing more cautiously, we realize that Avatar Korra is not really in the frame. The lady in a red outfit bears a facial resemblance to Avatar Korra,, and in the background of the video sequence, another female character bears a similar posture, skin tone and clothing style to Avatar Korra. In all, the model found a large number of features similar to those of Avatar Korra.

Figure 8 (c) illustrates a glitch in the system. Recall that in Figures 4, 5, 6 and 7 of success cases, we saw how the model, after training, can identify the Avatar even with different attire, hairstyle and clothing colour. However, in this present figure, the model was unable to classify Avatar Korra. One of the first reasons is that here, the Avatar wore an outfit that was never worn in the previous figures. Furthermore, the head protection that she wore did not allow the model to find at least a small number of the features that represent Avatar Korra. In short, the detection system failed because it reported not the target object while it was there. This figure case is a False Negative.

In Figure 8 (d), we observe a system failure when Avatar Korra is lying on the ground. The system fails to detect it in most cases during model training. The algorithm did not find many examples with this position, so when searching for Avatar Korra with this position, the system considered it as another object. In addition, there was the issue linked to her hairstyle, which in this case, was presented with her hair loose and spread.

Finally, Figure 8 (e) presents another glitch we observed during very fast-paced activities, for example, those involving a brawl or hand-to-hand combat.

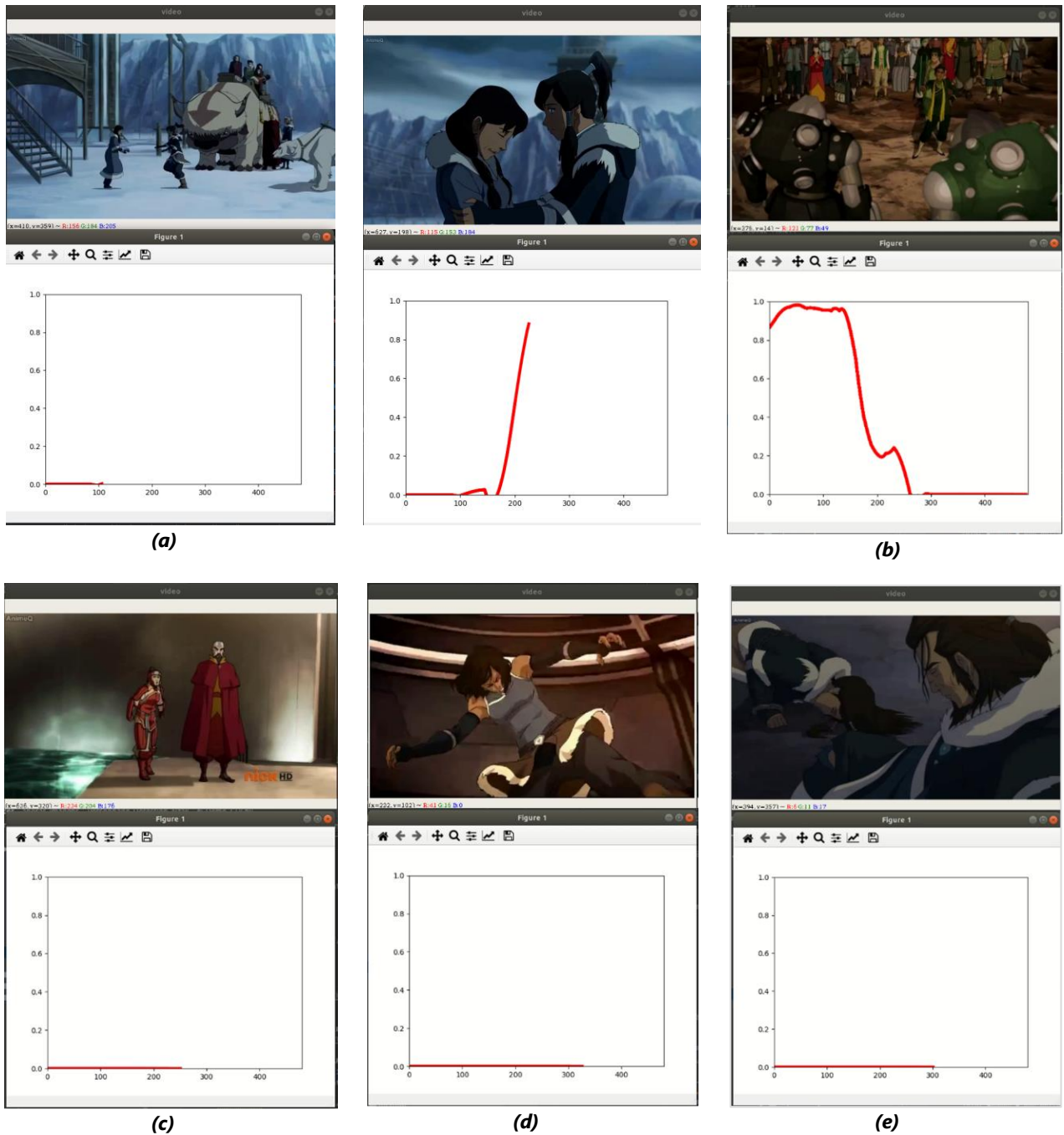

**Figure 8:** Failure case. (a) Difficulty in identifying Avatar Korra when she is distant and appears in small size of the video sequence. (b) Identification of Avatar Korra in the same video sequence in close-up. (c) False Positive 1. (d) False Positive 2. (e) False Positive 3. (f) Avatar Korra lying on the ground.

## 5. Conclusion

In this paper, we presented an approach from the Computer Vision branch based on supervised learning techniques for multimedia search. Our case study focused on both character classification and detection using validated dataset. The paper proposes a multi-instance and optimized learning model to find the set of points in the search space that best satisfies the objective function.

Considering the experiments performed, the experimental results obtained from the suggested model show positive results and potential implications for improving the efficiency of character detection using two artificial neural network structures for both the extraction and pattern recognition of the object to be classified. Considering the experiments performed and the results obtained, we conclude that the developed models can obtain a high success rate in detecting the character of interest (91%). Although the overall AUC value can be considered relatively weak, it should be considered that the model was able to obtain multiple situations in which the target object has been correctly detected. The trained model was also able to identify and classify it in multiple postures, and overall, the success cases exceeded the failure cases.

However, our model found some limitations, mainly in terms of recognition when the character was wearing certain clothing that did not present the characteristics that define it. However, limitations related to the case in which the head area was covered in more than 60% have been found. For future works, we propose future work to refine the accuracy of the proposed model and optimize the algorithm, applying it to the images captured by the CISP - Integrated Centre of Public Security of Angola, which is an opportunity to obtain an efficient improvement in combating crime.

**Funding:** This research received no external funding.
**Conflicts of Interest:** The authors declare no conflict of interest.
**Publisher's Note**: All claims expressed in this article are solely those of the authors and do not necessarily represent those of their affiliated organizations, or those of the publisher, the editors and the reviewers.

## References

[1]  Ahmad, M., Ahmed, I., & Adnan, A. (2019, October). Overhead view person detection using YOLO. In *2019 IEEE 10th annual ubiquitous computing, electronics & mobile communication conference (UEMCON)* (pp. 0627-0633). IEEE.

[2]  Amores, J. (2013). Multiple instance classification: Review, taxonomy and comparative study. *Artificial intelligence*, *201*, 81-105.

[3]  Bertolino, P., Foret, G., Pellerin, D., & Bertolino, P. (2001). Détection de personnes dans les vidéos pour leur immersion dans un espace virtuel. In *GRETSI, 18ème Colloque sur le Traitement du Signal et de l'Image*.

[4]  Carneiro, G., Chan, A. B., Moreno, P. J., & Vasconcelos, N. (2007). Supervised learning of semantic classes for image annotation and retrieval. *IEEE Transactions on pattern analysis and machine intelligence*, *29*(3), 394-410.

[5]  Dietterich, T. G., Lathrop, R. H., & Lozano-Pérez, T. (1997). Solving the multiple instance problems with axis-parallel rectangles. *Artificial intelligence*, *89*(1-2), 31-71.

[6]  Dutran (2019). Facebook archive/C3D. Available from https://github.com/facebookarchive/C3D, accessed on 27/06/2020.

[7]  *Dutran (Last modified Mar 20, 2017). C3D User Guide*. Retrieved from https://docs.google.com/document/d/1-QqZ3JHd76JfimY4QKqOojcEaf5g3JS0lNh-FHTxLag/edit

[8]  Dhiman, C., & Vishwakarma, D. K. (2019). A Robust Framework for Abnormal Human Action Recognition Using $\boldsymbol{\mathcal{R}}$-Transform and Zernike Moments in Depth Videos. *IEEE Sensors Journal*, *19*(13), 5195-5203.

[9]  Foresti, G. L., Micheloni, C., & Piciarelli, C. (2005). Detecting moving people in video streams. *Pattern Recognition Letters*, *26*(14), 2232-2243.

[10]  Jagadeesh, B., & Patil, C. M. (2016, May). Video based action detection and recognition humans using optical flow and SVM classifier. In *2016 IEEE International Conference on Recent Trends in Electronics, Information & Communication Technology (RTEICT)* (pp. 1761-1765). IEEE.

[11]  Kim, J., Park, Y., Choi, K. P., Lee, J., Jeon, S., & Park, J. (2017, September). Dynamic frame resizing with convolutional neural network for efficient video compression. In *Applications of Digital Image Processing XL* (Vol. 10396, pp. 343-355). SPIE.

[12]  Krizhevsky, A., Sutskever, I., & Hinton, G. E. (2017). Imagenet classification with deep convolutional neural networks. *Communications of the ACM*, *60*(6), 84-90.

[13]  Lin, T., Yu, Z., Hu, H., Xu, Y., & Chen, C. W. (2023). Interventional bag multi-instance learning on whole-slide pathological images. In *Proceedings of the IEEE/CVF Conference on Computer Vision and Pattern Recognition* (pp. 19830-19839).

[14]  Quddus, A., Fieguth, P., & Basir, O. (2006, January). Adaboost and support vector machines for white matter lesion segmentation in MR images. In *2005 IEEE Engineering in Medicine and Biology 27th Annual Conference* (pp. 463-466). IEEE.

[15]  Quddus, A., Fieguth, P., & Basir, O. (2005). Adaboost and Support Vector Machines for White Matter Lesion Segmentation in MR Images. *In 2005 IEEE Engineering in Medicine and Biology 27th Annual Conference*, Shanghai.

[16]  Sultani, W., Chen, C., & Shah, M. (2018). Real-world anomaly detection in surveillance videos. In *Proceedings of the IEEE conference on computer vision and pattern recognition* (pp. 6479-6488).

[17]  Wikipedia (2019). *Multiple Instance Learning*. Accessed on 15/11/2022. https://en.wikipedia.org/wiki/Multiple_instance_learning.

[18]  Watch Movies Online, accessed 29 November 2019. available from :http://www.nick.co.uk/shows/legendofkorra/videos/airbending-basics/0sxstw,

[19]  Yang, C., Dong, M., & Hua, J. (2006, June). Region-based image annotation using asymmetrical support vector machine-based multiple-instance learning. In *2006 IEEE Computer Society Conference on Computer Vision and Pattern Recognition (CVPR'06)* (Vol. 2, pp. 2057-2063). IEEE.